\def\year{2019}\relax
\documentclass[letterpaper]{article} 
\usepackage{aaai19}  
\usepackage{times}  
\usepackage{helvet}  
\usepackage{courier}  
\usepackage{url}  
\usepackage{graphicx}  
\usepackage{todonotes}
\usepackage{amsfonts} 

\usepackage{tabularx}
\newcolumntype{L}[1]{>{\raggedright\arraybackslash}p{#1}}

\begin{document}

\title{A Hierarchical Multi-task Approach for Learning Embeddings from Semantic Tasks}
\author{Victor Sanh\textsuperscript{1}, Thomas Wolf\textsuperscript{1}, Sebastian Ruder\textsuperscript{2,3}\\
\textsuperscript{1}{Hugging Face, 20 Jay Street, Brooklyn, New York, United States}\\
\textsuperscript{2}{Insight Research Centre, National University of Ireland, Galway, Ireland}\\
\textsuperscript{3}{Aylien Ltd., 2 Harmony Court, Harmony Row, Dublin, Ireland}\\
\texttt{\{victor, thomas\}@huggingface.co}, \texttt{sebastian@ruder.io}
}

\maketitle

\begin{abstract}
    Much effort has been devoted to evaluate whether multi-task learning can be leveraged to learn rich representations that can be used in various Natural Language Processing (NLP) down-stream applications. However, there is still a lack of understanding of the settings in which multi-task learning has a significant effect. In this work, we introduce a hierarchical model trained in a multi-task learning setup on a set of carefully selected semantic tasks. The model is trained in a hierarchical fashion to introduce an inductive bias by supervising a set of low level tasks at the bottom layers of the model and more complex tasks at the top layers of the model. This model achieves state-of-the-art results on a number of tasks, namely Named Entity Recognition, Entity Mention Detection and Relation Extraction without hand-engineered features or external NLP tools like syntactic parsers. The hierarchical training supervision induces a set of shared semantic representations at lower layers of the model. We show that as we move from the bottom to the top layers of the model, the hidden states of the layers tend to represent more complex semantic information.
\end{abstract}

\section{Introduction}

Recent Natural Language Processing (NLP) models heavily rely on rich distributed representations (typically word or sentence embeddings) to achieve good performance. One example are  so-called ``universal representations'' \cite{Conneau17} which are expected to encode a varied set of linguistic features, transferable to many NLP tasks. This kind of rich word or sentence embeddings can be learned by leveraging the training signal from different tasks in a multi-task setting. It is known that a model trained in a multi-task framework can take advantage of inductive transfer between the tasks, achieving a better generalization performance \cite{Caruana1993}. Recent works in sentence embeddings \cite{Subramanian2018,Jernite2017} indicate that complementary aspects of the sentence (e.g. syntax, sentence length, word order) should be encoded in order for the model to produce sentence embeddings that are able to generalize over a wide range of tasks. Complementary aspects in representations can be naturally encoded by training a model on a set of diverse tasks, such as, machine translation, sentiment classification or natural language inference. Although, (i) the selection of this diverse set of tasks, as well as, (ii) the control of the interactions between them are of great importance, a deeper understanding of (i) and (ii) is missing as highlighted in the literature~\cite{Caruana1997,Mitchell80theneed,Ruder2017}.
This work explores this line of research by combining, in a single model, four fundamental semantic NLP tasks: Named Entity Recognition, Entity Mention Detection (also sometimes referred as mention detection), Coreference Resolution and Relation Extraction. This selection of tasks is motivated by the inter-dependencies these tasks share. In Table~\ref{tab:motivating_examples}, we give three simple examples to exemplify the reasons why these tasks should benefit from each other. For instance, in the last example knowing that \textit{the company} and \textit{Dell} are referring to the same real world entity, \textit{Dell} is more likely to be an organization than a person. 

\begin{table}
    \caption{A few examples motivating our selection of tasks. \newline Abbreviations: CR: Coreference Resolution, RE: Relation Extraction, EMD: Entity Mention Detection, NER: Named Entity Recognition, X $\rightsquigarrow$ Y: X is more likely to be Y.}
    \label{tab:motivating_examples}
    \centering
    \resizebox{\columnwidth}{!}{
    \begin{tabular}{L{0.43\columnwidth}L{0.35\columnwidth}L{0.41\columnwidth}}
        \hline
        Example & Predictions on one task... & \hspace{0pt}...can help disambiguate other tasks \\
        \hline
        \hline
        \textit{X works for Y} & RE: \{\textit{work, X, Y}\} & \textit{X} $\rightsquigarrow$ Person (EMD)\newline \textit{Y} $\rightsquigarrow$  Organization or Person (NER)\\
        \hline
        \textit{I love Melbourne. I've lived three years in this city.} & CR: (\textit{Melbourne, this city}) \newline RE: \{\textit{live, I, this city}\} & \textit{Melbourne} $\rightsquigarrow$ Location (EMD/NER)\\
        \hline
        \textit{Dell announced a \$500M net loss. The company is near bankruptcy.} & CR: (\textit{Dell, The company}) & \textit{Dell} $\rightsquigarrow$ Organization (EMD/NER) \\
        \hline
    \end{tabular}
    }
\end{table}

Several prior works \cite{Yang2016,BingelS17} avoid the question of the linguistic hierarchies between NLP tasks. We argue that some tasks (so-called ``low level'' tasks) are simple and require a limited amount of modification to the input of the model while other tasks (so-called ``higher level'' tasks) require a deeper processing of the inputs and likely a more complex architecture. Following \cite{Hashimoto2017,Sogaard2016}, we therefore introduce a hierarchy between the tasks so that low level tasks are supervised at lower levels of the architecture while keeping more complex interactions at deeper layers.
Unlike previous works \cite{Li2014,Miwa2016}, our whole model can be trained end-to-end without any external linguistic tools or hand-engineered features while giving stronger results on both Relation Extraction and Entity Mention Detection.

Our main contributions are the following:
(1) we propose a multi-task architecture combining four different tasks that have not been explored together to the best of our knowledge. This architecture uses neural networks and does not involve external linguistic tools or hand-engineered features. We also propose a new sampling strategy for multi-task learning, \textit{proportional sampling}.
(2) We show that this architecture can lead to state-of-the art results on several tasks e.g. Named Entity Recognition, Relation Extraction and Entity Mention Detection while using simple models for each of these tasks. This suggests that the information encoded in the embeddings is rich and covers a variety of linguistic phenomena.
(3) We study and give insights on the influence of multi-task learning on (i) the speed of training and (ii) the type of biases learned in the hierarchical model.

\section{Model}

\begin{figure}
    \includegraphics[width=\linewidth]{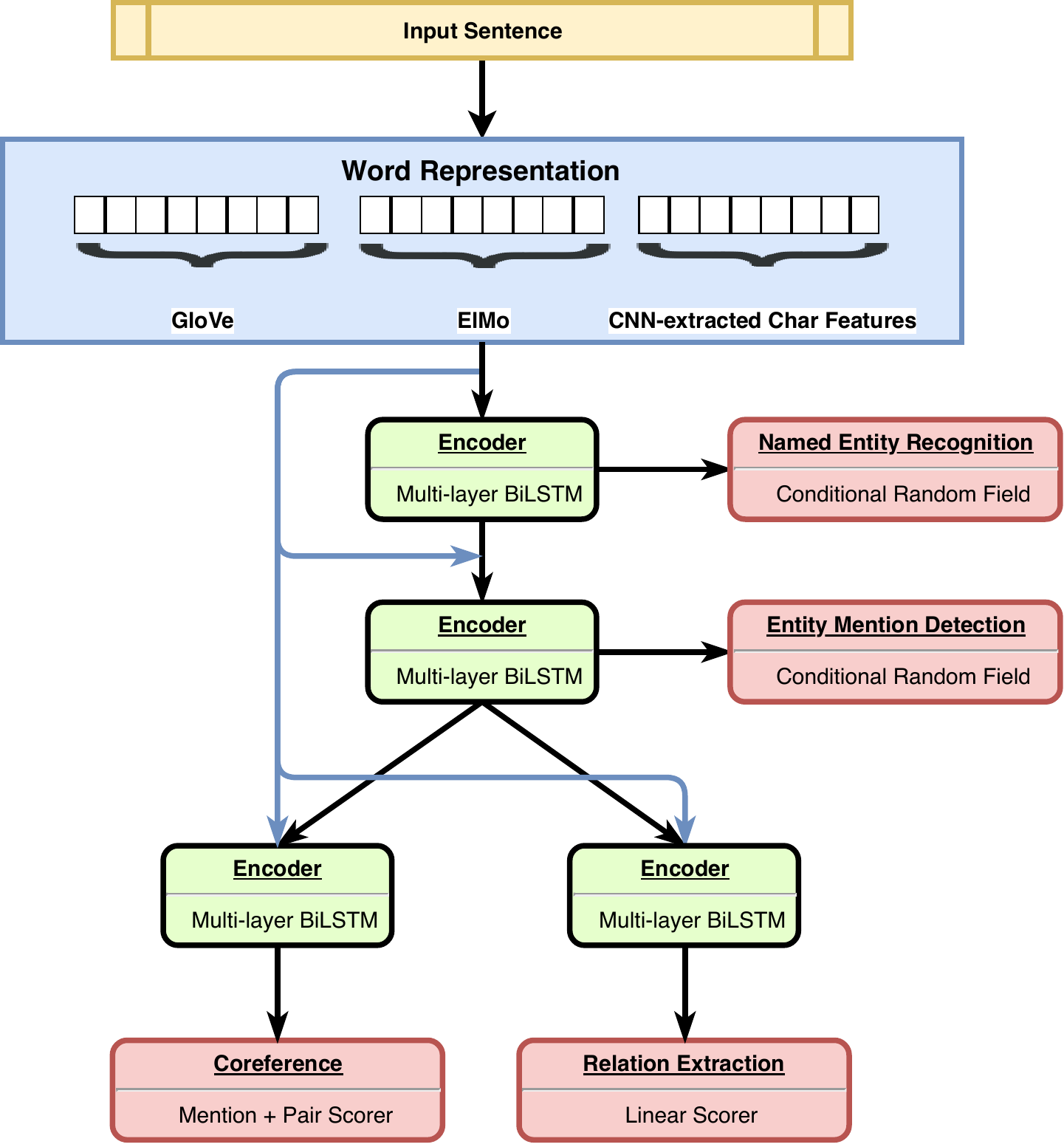}
    \caption{Diagram of the model architecture}
    \label{diag:model_architecture}%
\end{figure}

In this section, we describe our model beginning at the lower levels of the architecture and ascending to the top layers. Our model introduces a hierarchical inductive bias between the tasks by supervising low-level tasks (that are assumed to require less knowledge and language understanding) at the bottom layers of the model architecture and supervising higher-level tasks at higher layers. The architecture of the model is shown in Figure~\ref{diag:model_architecture}. Following \cite{Hashimoto2017}, we use shortcut connections so that top layers can have access to bottom layer representations.

\subsection{Words embeddings}

Our model encodes words $w_t$ of an input sentence $s = (w_1, w_2, ..., w_n)$ as a combination of three different types of embeddings. We denote the concatenation of the these three embeddings $g_{e}$.\\ 
\underline{Pre-trained word embeddings}: We use GloVe \cite{pennington2014glove} pre-trained word level embeddings. These embeddings are fine-tuned during training.\\
\underline{Pre-trained contextual word embeddings}: We also use contextualized ELMo embeddings \cite{Peters2018}. These word embeddings differ from GloVe word embeddings in that each token is represented by a vector that is a function of the whole sentence (a word can thus have different representations depending on the sentence it is extracted from). These representations are given by the hidden states of a bi-directional language model. ELMo embeddings have been shown to give state-of-the-art results in multiple NLP tasks \cite{Peters2018}.\\
\underline{Character-level word embeddings}: Following \cite{Chiu2015,Lample2016}, we use character-level word embeddings to extract character-level features.
Specifically, we use a convolutional neural network (CNN) (followed by a max pooling layer) for the ease of training since Recurrent Neural Network-based encodings do not significantly outperform CNNs while being computationally more expensive to train \cite{Ling2015}.

\subsection{Named Entity Recognition (NER)}

The first layers of our model are supervised by Named Entity Recognition labels.  NER aims to identify mentions of named entities in a sequence and classify them into pre-defined categories. In accordance with previous work \cite{Chiu2015,Lample2016} the tagging module contains an RNN-based encoding layer followed by a sequence tagging module based on a conditional random field \cite{Lafferty2001}. We use multi-layer bi-directional LSTMs (Long Short-Term Memory) as encoders.
The encoders take as input the concatenated word embeddings $g_{e}$ and produce (sequence) embeddings $g_{ner}$. Specifically, $g_{ner}$ are the concatentation of the backward and forward hidden states of the top layer of the biLSTMs, which are then fed to the sequence tagging layer.

We adopt the BILOU (Beginning, Inside, Last, Outside, Unit) tagging scheme. The tagging decisions are modeled using a CRF, which explicitly reasons about interactions between neighbour tokens tags.

\subsection{Entity Mention Detection (EMD)}

A second group of layers of our model are supervised using Entity Mention Detection labels. EMD is similar in spirit to NER but more general as it aims to identify all the mentions related to a real life entity, whereas NER only focuses on the named entities. Let us consider an example: \textit{$[$The men$]_{PERS}$ held on $[$the sinking vessel$]_{VEH}$ until $[$the passenger ship$]_{VEH}$ was able to reach them from $[$Corsica$]_{GPE}$.} Here, NER annotations would only tag \textit{Corsica}, while EMD requires a deeper understanding of the entities in the sentence. 

We formulate Mention Detection as a sequence tagging task using a BILOU scheme. We use a multi-layer biLSTM followed by a CRF tagging layer. We adopt shortcut connections so that each layer can build on top of the representations extracted by the lower layers in a hierarchical fashion. The encoder thus takes as input the concatenation of the lower layer representations $[g_{e}, g_{ner}]$ and outputs sequence embeddings denoted by $g_{emd}$.

To be able to compare our results with previous works \cite{Bekoulis2018,Miwa2016,Katiyar2017} on EMD, we identify the head of the entity mention rather than the whole mention.

\subsection{Coreference Resolution (CR)}

Ascending one layer higher in our model, CR is the task of identifying mentions that are referring to the same real life entity and cluster them together (typically at the level of a few sentences). For instance, in the example \textit{My mom tasted the cake. She said it was delicious}, there are two clusters: (My mom, She) and (the cake, it). CR is thus a task which requires a form of semantic representation to cluster the mentions pointing to the same entity.
 
We use the model proposed in \cite{Lee2017}. This model considers all the spans in a document as potential mentions and learns to distinguish the candidate coreferent mentions from the other spans using a mention scorer to prune the number of mentions. The output of the mention scorer is fed to a mention pair scorer, which decides whether identified candidates mentions are coreferent. The main elements introduced in \cite{Lee2017} are the use of span embeddings to combine context-dependent boundary representation and an attention mechanism over the span to point to the mention's head. This model is trained fully end-to-end without relying on external parser pre-processing.

The encoder takes as input the concatenation of $[g_{e}, g_{emd}]$ and outputs representations denoted as $g_{cr}$, which are then fed to the mention pair scorer.

\subsection{Relation Extraction (RE)}

The last supervision of our model is given by Relation Extraction (RE). RE aims at identifying semantic relational structures between entity mentions in unstructured text. Traditional systems treat this task as two pipelined tasks: (i) identifying mentions and (ii) classifying the relations between identified mentions. We use the Joint Resolution Model proposed by \citeauthor{Bekoulis2018} \shortcite{Bekoulis2018} in which the selection of the mentions and classification of the relation between these mentions are performed jointly. Following previous work \cite{Li2014,Katiyar2017,Bekoulis2018}, we only consider relations between the last token of the head mentions involved in the relation. Redundant relations are therefore not classified.

The RE encoder is a multi-layer BiLSTM which takes as input $[g_{e}, g_{emd}]$ and outputs a representation denoted $g_{re}$. These contextualized representations are fed to a feedforward neural network. More specifically, considering two token's contextualized representations, $g_i$ and $g_j$, both of size $\mathbb{R}^{l}$, we compute a vector score:
\begin{equation}
t(w_i, w_j) = V \phi (U g_j + W g_i + b)
\end{equation}
where $U \in \mathbb{R}^{d \times l}$, $W \in \mathbb{R}^{d \times l}$, $b \in \mathbb{R}^{d}$, and $V \in \mathbb{R}^{r \times d}$ are learned transformation weights, $l$ is the size of the embeddings output by the encoder, $d$ is the size of the hidden layer of the feedforward network, $r$ is the number of possible relations, and $\phi$ is a non-linear activation function. The relation probabilities are then estimated as $p = \sigma (t(w_i, w_j)) \in \mathbb{R}^{r}$ where $p_k$ ($1 \leq k \leq r$) is the probability that token $w_i$ and token $w_j$ are respectively labeled as \textit{ARG1} and \textit{ARG2} in a relation of type $k$. The model predictions are computed by thresholding estimated probabilities. The parameters of the model $V$, $U$, $W$, and $b$ are trained by minimizing a cross-entropy loss.

In this formulation, a mention may be involved in several relations at the same time (for instance being the \textit{ARG1} and the \textit{ARG2} in two respective relations), which can occur in real life examples. If we replaced the \textit{sigmoid} function by a \textit{softmax} function, this is not possible.

In the model, the CR and RE modules are both on the same level. We did not find it helpful to introduce a hierarchical relation between these two tasks as they both rely on deeper semantic modeling, i.e. both trying to link mentions.

\section{Experiment setting}

\subsection{Datasets and evaluation metrics}

We use labeled data from different sources to train and evaluate our model. For NER, we use the English portion of OntoNotes 5.0 \cite{OntoNotes2}. Following \citeauthor{Strubell2017} \shortcite{Strubell2017}, we use the same data split as used for coreference resolution in the CoNLL-2012 shared task \cite{Pradhan2012Conll2012ST}. We report the performance on NER using span level $F_{1}$ score on the test set. The dataset covers a large set of document types (including telephone conversations, web text, broadcast news and translated documents), and a diverse set of 18 entity types (including PERSON, NORP, FACILITY, ORGANIZATION, GPE). Statistics of the corpus are detailed in Table~\ref{ontonotes_statistics}. We also report performance on more commonly used CoNLL2003 NER dataset.

For CR, EMD and RE, we use the Automatic Content Extraction (ACE) program ACE05 corpus \cite{ACE05}. The ACE05 corpus is one of the largest corpus annotated with CR, EMD and RE making it a compelling dataset for multi-task learning. Mention tags in ACE05 cover 7 types of entities such as Person, Organization, or Geographical Entities. For each entity, both the mention boundaries and the head spans are annotated. ACE05 also introduces 6 relation types (including Organization-Affiliation (ORG-AFF), GEN-Affiliation (GEN-AFF), and Part-Whole (PART-WHOLE)). We use the same data splits as previous work \cite{Li2014,Miwa2016,Katiyar2017} for both RE and EMD and report $F_{1}$-scores, Precision, and Recall. We consider \textit{an entity mention} correct if the model correctly predicted both the mention's head and its type. We consider \textit{a relation} correct if the model correctly predicted the heads of the two arguments and the relation type.

For CR, we use different splits to be able to compare to previous work \cite{Bansal2012,Durrett2014}. These splits (introduced in \cite{Rahman2009}) use the whole ACE05 dataset leaving 117 documents for test while having 482 documents for training (as in \cite{Bansal2012}, we randomly split the training into a 70/30 ratio to form a validation set). We evaluate coreference on both splits. We compare all  coreference systems using the commonly used metrics: MUC, B3, CEAFe (CEAF$\phi_4$) as well as the average $F_{1}$ of the three metrics as computed by the official CoNLL-2012 scorer. Note that \citeauthor{Durrett2014} make use of external NLP tools including an automatic parser \cite{Durrett13easyvictories}.

We compare our model to several previous systems that have driven substantial improvements over the past few years both using graphical models or neural-net-based models. These are the strongest baselines to the best of our knowledge.

\begin{table}
  \caption{Data statistics}
  \label{ontonotes_statistics}
  \centering
  \begin{tabular}{lccc}
    \hline
    OntoNotes & Train & Dev & Test \\
    \hline
    Documents & 2,483 & 319 & 322 \\
    Sentences & 59,924 & 8,529 & 8,262 \\
    Named Entities & 81,828 & 11,066 & 11,257 \\
    Tokens & 1,088,503 & 147,724 & 152,728 \\
    \hline
    ACE05 & Train & Dev & Test \\
    \hline
    Documents & 351 & 80 & 80 \\
    Sentences & 7,273 & 1,765 & 1,535 \\
    Mentions & 26,470 & 6,421 & 1,535 \\
    Relations & 4,779 & 1,179 & 1,147 \\
    \hline
  \end{tabular}
\end{table}

\subsection{Training Details}
\label{subsec:training}

\citeauthor{Subramanian2018} \shortcite{Subramanian2018} observe that there is no clear consensus on how to correctly train a multi-task model. Specifically, there remain many open questions such as ``when should the training schedule switch from one task to another task?'' or ``should each task be weighted equally?'' One of the main issues that arises when training a multi-task model is \textit{catastrophic forgetting} \cite{French1999} where the model abruptly forgets part of the knowledge related to a previously learned task as a new task is learned. This phenomenon is especially present when multiple tasks are trained sequentially.

We selected the simple yet effective training method described in \cite{Sogaard2016,Ruder2017c}: after each parameter update, a task is randomly selected and a batch of the dataset attached to this task is also sampled at random to train the model. This process is repeated until convergence (the validation metrics do not improve anymore). We tested both uniform and proportional sampling and found that proportional sampling performs better both in terms of performance and speed of convergence. In proportional sampling, the probability of sampling a task is proportional to the relative size of each dataset compared to the cumulative size of all the datasets. Note that unlike \cite{Subramanian2018}, the updates for a particular task affect the layers associated with this task and all the layers below but not the layers above.

\section{Results and Discussion}

\begin{table*}
\centering
\caption{Results: Baselines and ablation study on the tasks. 
\textit{GM} means that the same coreference module uses gold mentions at evaluation time and that we used the splits introduced in \cite{Rahman2009}. Otherwise, we use for coreference the same splits as for EMD and RE (351/80/80). For coreference, figures that are comparable with \cite{Durrett2014} are tagged with an *.}
\label{table:ablation_task}
\resizebox{2\columnwidth}{!}{%
    \begin{tabular}{|l l|| c|c|c || c|c|c || c|c|c || c|c|c|c |}
       \hline
        & & \multicolumn{3}{c||}{NER} & \multicolumn{3}{c||}{EMD} & \multicolumn{3}{c||}{RE} & \multicolumn{4}{c|}{CR} \\
       \hline
       Setup & Model & 
            $P$ & $R$ & $F_{1}$ &  
            $P$ & $R$ & $F_{1}$ & 
            $P$ & $R$ & $F_{1}$ &
            MUC & B3 & Ceafe & Avg. $F_{1}$ \\
       \hline
       \hline
       & \cite{Strubell2017} & - & - & 86.99
                & - & - & - 
                & - & - & - 
                & - & - & - & -  \\
       & \cite{Katiyar2017} & - & - & - 
                & 84.0 & 81.3 & 82.6
                & 57.9 & 54.0 & 55.9
                & - & - & - & -  \\
        & \cite{Miwa2016} & - & - & - 
                & 82.9 & 83.9 & 83.4
                & 57.2 & 54.0 & 55.6
                & - & - & - & -  \\
        & \cite{Li2014} & - & - & -  
                & 85.2 & 76.9 & 80.8
                & 68.9 & 41.9 & 52.1
                & - & - & - & -  \\
        & \cite{Durrett2014} & - & - & -  
                & - & - & -  
                & - & - & -  
                & 81.03* & 74.89* & 72.56* & 76.16* \\
        & \cite{Bansal2012} & - & - & -  
                & - & - & -  
                & - & - & -  
                & 70.2* & 72.5* & - & - \\
       \hline
       \hline
       (A) & Full Model 
                & 87.52 & 87.21 & 87.36 
                & 85.68 & 85.69 & 85.69
                & 68.53 & 54.48 & 61.30
                & 73.89 & 61.34 & 59.11 & 64.78 \\
        (A-GM) & Full Model - GM
                & 87.12 & 87.09 & 87.10 
                & 87.15 & 87.33 & \textbf{87.24}
                & 70.40 & 56.40 & \textbf{62.69}
                & 82.49* & 67.64* & 60.75* & \textbf{70.29}* \\
       \hline
       (B) & NER 
            & 87.24 & 87.00 & 87.12
            & - & - & -
            & - & - & -
            & - & - & - & - \\
        (C) & EMD 
            & - & - & - 
            & 87.03 & 85.27 & 86.14
            & - & - & -
            & - & - & - & - \\
        (D) & RE 
            & - & - & - 
            & - & - & -
            & 60.47 & 52.14 & 55.99
            & - & - & - & - \\
        (E) & CR 
            & - & - & - 
            & - & - & -
            & - & - & -
            & 74.80 & 62.63 & 59.59 & 65.67 \\
        (E-GM) & CR - GM 
            & - & - & - 
            & - & - & -
            & - & - & -
            & 81.78* & 66.42* & 59.93* & 69.38* \\
        \hline
        (F) & NER + EMD 
            & 87.50 & 86.32 & 86.91
            & 86.54 & 85.49 & 86.02
            & - & - & -
            & - & - & - & - \\
        (G) & EMD + RE 
            & - & - & -
            & 85.58 & 85.38 & 85.50
            & 68.66 & 54.05 & 60.49
            & - & - & - & - \\
        (H) & EMD + CR 
            & - & - & -
            & 85.84 & 85.46 & 85.65
            & - & - & -
            & 72.67 & 59.05 & 57.33 & 63.02 \\
        \hline
        (I) & NER + EMD + RE 
            & 87.37 & 87.65 & \textbf{87.51} 
            & 86.63 & 85.90 & 86.26
            & 65.57 & 55.62 & 60.18
            & - & - & - & - \\
        (J) & NER + EMD + CR 
            & 87.67 & 87.34 & 87.50 
            & 85.89 & 85.86 & 85.87
            & - & - & -
            & 75.73 & 62.92 & 61.24 & \textbf{66.64} \\
        \hline
        (K) & EMD + NER 
            & 85.67 & 87.19 & 86.43 
            & 85.62 & 84.76 & 85.19 
            & - & - & -
            & - & - & - & - \\
        (L) & EMD + NER + RE + CR 
            & 85.78 & 86.66 & 86.21 
            & 85.24 & 85.05 & 85.14 
            & 63.32 & 55.54 & 59.17 
            & 73.29 & 60.37 & 58.86 & 64.17 \\
        \hline
    \end{tabular}%
}
\end{table*}

\subsection{Overall Performance}

In this section, we present our main results on each task and dataset. The hierarchical model and multi-task learning framework presented in this work achieved state-of-the-art results on three tasks, namely NER (+0.52), EMD (+3.8) and RE (+6.8). Table~\ref{table:ablation_task} summarizes the results and introduces each setups' abbreviation (alphabetical letters). In the following subsections, we highlight a few useful observations.

To be able to compare our work on CR with the various baselines, we report results using different settings and splits. More precisely, \textit{GM} indicates that gold mentions were used for evaluation and that coreference was trained using the ACE05 splits introduced in \cite{Rahman2009}.

Using gold mentions is impossible in real settings so we also relax this condition leading to a more challenging task in which we make no use of external tools or metadata (such as speaker ID used by some systems \cite{Clark2015}). Comparing setups A and A-GM shows how the supervision from one module (e.g. CR) can flow through the entire architecture and impact other tasks' performance: RE's $F_{1}$ score drops by $\sim$1 point on A. Note that the GM setup impacts the training exit condition (the validation metrics stop improving) and the evaluation metrics (it is well known that using gold mentions at evaluation time improves CR's performance).
Similarly, the A-GM setup leads to the state-of-the-art on EMD and RE. It increases the $F_{1}$ by $\sim$1.5 points for EMD and $\sim$1 point for RE (A vs. A-GM). This suggests that having different type of information on different sentences brings richer information than having multiple types of information on the same sentences (setup A-CoNLL2012 -see Table~\ref{table:canocical}- supports this claim as CR trained on another dataset leads to comparable performance on the three other tasks).

To analyze which components of the model are driving the improvements and understand the interactions between the different tasks and modules, we performed an ablation study summarized in the following paragraphs and on Tables~\ref{table:ablation_task} and~\ref{table:ablation_embeddings}.

\subsubsection*{Single Task vs. Full Models}

The largest difference between a single-task and a multi-task setup is observed on the RE task (A vs. D on Table~\ref{table:ablation_task}), while the results on NER are similar in multi-task and single-task setups (B vs. A \& A-GM). This further highlights how the RE module can be sensitive to information learned from other tasks. Results on EMD are in the middle, with the single task setup giving higher score than a multi-task-setup except for A-GM and I. More surprisingly, CR can give slightly better results when being single-task-trained (A vs. E).

\subsubsection*{Progressively adding tasks}

To better understand the contribution of each module, we vary the number of tasks in our training setup. The experiments show that training using RE helps both for NER and for EMD. Adding RE supervision leads to an increase of $\sim$1 point on NER while boosting both precision and recall on EMD (F vs. I). CR and RE can help NER as shown by comparing setups A and F: recall and $F_{1}$ for NER are $\sim$1 point stronger, while the impact on EMD is negative. Finally, training using CR supervision boosts NER (F vs. J) by increasing NER's recall while lowering EMD's precision and $F_{1}$. In other words the information flowing along the hierarchical model (e.g. stacking of encoders and shortcut connections) enables higher levels' supervision to train lower levels to learn better representations. More generally, whenever the task RE is combined with another task, it always increases the $F_{1}$ score (most of the improvement coming from the precision) by 2--6 $F_{1}$ points. 

\subsubsection*{Experimenting with the hierarchy order}

Comparing setups F vs. K and A vs. L in which we switched NER and EMD, provides evidence for the hierarchical relation between NER and EMD: supervising EMD at a lower level than NER is detrimental to the overall performance. This supports our intuition that the hierarchy should follow the intrinsic difficulty of the tasks.

\subsubsection*{Comparison to other canonical datasets}

We also compare our model on two other canonical datasets for NER (CoNLL-2003) and CR (CoNLL-2012). Details are reported in Table~\ref{table:canocical}. We did not tune hyperparameters, keeping the same hyperparameters as used in the previous experiments. We reach performance comparable to previous work and the other tasks, demonstrating that our improvements are not dataset-dependent.

\begin{table}
\centering
\caption{Comparison to other canonical datasets on NER (CoNLL-2003) and coreference (CoNLL-2012). A-CoNLL: train A-RS-GM using CoNLL-2003 for NER; A-CoNLL-2012: train A using CoNLL-2012 for coreference.}
\label{table:canocical}
\resizebox{0.8\columnwidth}{!}{%
    \begin{tabular}{|l | c  | c |}
       \hline
        Model & 
            NER ($F_{1}$) & CR ($F_{1}$) \\
       \hline
        \citeauthor{Lample2016} \shortcite{Lample2016}
            & 90.94 & - \\        
        \citeauthor{Strubell2017} \shortcite{Strubell2017}
            & 90.54 & - \\
        \citeauthor{Peters2018} \shortcite{Peters2018}
            & 92.22 & - \\
        (A-CoNLL-2003) & 91.63 & 70.14 \\
        \hline
         \citeauthor{Durrett2014} \shortcite{Durrett2014} & - & 61.71 \\ 
         \citeauthor{Lee2017} \shortcite{Lee2017} (single) & - & 67.2 \\
         \citeauthor{Lee2017} \shortcite{Lee2017} (ensemble) & - & 68.8 \\         
        (A-CoNLL-2012) & 86.90 & 62.48 \\
        \hline
    \end{tabular}%
}
\end{table}

\subsubsection*{Effect of the embeddings}

We perform an ablation study on the words and character embeddings $g_{e}$. Results are reported in Table~\ref{table:ablation_embeddings}. As expected, contextualized ELMo embeddings have a noticeable effect on each metrics. Removing ELMo leads to a $\sim$4 $F_{1}$ points drop on each task. Furthermore, character-level embeddings, which are sensitive to morphological features such as prefixes and suffixes and capitalization, also have a strong impact, in particular on NER, RE and CR. Removing character-level embeddings does not affect EMD suggesting that the EMD module can compensate for this information. The main improvements on the CR task stem from the increase in B3 and Ceafe metrics. Note that the strong effect of removing a type of embedding is also a consequence of using shortcut connections: removing an embedding has a direct impact on the input to each task's module.

\begin{table}
\centering
\caption{Ablation study on the embeddings. We remove one by one the embeddings on the first layer of the best performing model (A-RS-GM).}
\label{table:ablation_embeddings}
\resizebox{\columnwidth}{!}{%
    \begin{tabular}{|l || c | c | c | c |}
       \hline
        Model & 
            NER ($F_{1}$) &  EMD ($F_{1}$) & RE ($F_{1}$) & CR ($F_{1}$) \\
       \hline
       \hline
        Glove + Char. embds + ELMo 
                & 87.10 
                & 87.24
                & 62.69
                & 70.29 \\
       \hline
        Glove + Char. embds 
            & 84.33
            & 83.13 
            & 57.47 
            & 66.44 \\
        Glove 
            & 79.81
            & 83.00 
            & 53.77 
            & 64.26 \\            
        \hline
    \end{tabular}%
}
\end{table}

\subsection{What did the embeddings and encoders learn?}

High scores on a specific task suggest that the representations learned by the encoders (and the embeddings) have somehow managed to capture relevant linguistic or statistical features for the task. However, using complex architectures makes it difficult to understand what is actually encoded in the embeddings and hidden states and what type of linguistic information, if any, is being used by the model to make a particular prediction. To further understand our architecture, we analyze the inductive biases encoded in the embeddings and hidden states of the various layers. We follow \citeauthor{Conneau2018} \shortcite{Conneau2018} who introduced 10 different probing tasks\footnote{A probing task is a classification task that focuses on a well identified linguistic property.} to analyze the quality of sentence embeddings. These tasks aim at evaluating a wide set of linguistic properties from surface information, to syntactic information through semantic information.

We use a simple logistic regression classifier, which takes the sentence embeddings as inputs and predicts the linguistic property. We study both the word embeddings ($g_{e}$) and the hidden state representations (biLSTM encoders) specific to each module in our model. The sentence embedding of an input sequence of length $L$ is computed from the $L$ hidden states of an encoder by taking the maximum value over each dimension of the last layer activations as done in \cite{Conneau17}.
Sentence embeddings are obtained from word and character-level embeddings by max-pooling over a sentence's words. Averaging word embeddings is known to be a strong baseline for sentence embeddings \cite{Arora2017asimple} and we also report the results of this simpler procedure in Table~\ref{table:senteval}.

\subsubsection*{Results}

We compare our results with two baselines from \cite{Conneau2018}: bag-of-words computed from FastText embeddings and SkipThought sentence embeddings \cite{kiros2015skip}. We compare the base word embeddings $g_{e}$ of our model with the first baseline and the output of the task-specific encoders to the second baseline. A first observation is that the word embeddings already encode rich representations, having an accuracy higher than 70\% on seven of the ten probing tasks. We suspect that shortcut connections are key to this good performances by allowing high level tasks to encode rich representations. The good performance on \textit{Bigram Shift} (compared to BoV-FastText: +38.8) likely comes from the use of ELMo embeddings which are sensitive to word order. The same argument may also explain the strong performance on \textit{Sentence Length}.

There are significant discrepancies between the results of the word embeddings $g_{e}$ and the encoder representations, indicating that the learned linguistic features are different between these two types of embeddings. Averaging the base embeddings surpasses encoder embeddings on almost all the probing tasks (except \textit{Coordination Inversion}). The difference is particularly high on the \textit{Word Content} task in which the results of the encoders embeddings barely rise above 11.0, indicating that the ability to recover a specific word is not a useful feature for our four semantic tasks.

The performance of the encoder representation is stronger on semantic probing tasks, compared to the low signal for surface and syntatic tasks. The only exception is the \textit{Sentence Length} which suggest that this linguistic aspect is naturally encoded. The performances of the NER and EMD encoders are generally in the same range supporting the fact that these two tasks are similar in nature. Finally, we observe that the highest scores for encoder representations always stem from the coreference encoder suggesting that CR is both the highest level task and that the CR module requires rich and diverse representations to make a decision.

\begin{table*}
\centering
\caption{SentEval Probing task accuracies. Classification is performed using a simple logistic regression enabling fair evaluation of the richness of a sentence embedding. We report two baselines from \citeauthor{Conneau2018}.}
\label{table:senteval}
\resizebox{2\columnwidth}{!}{%
    \begin{tabular}{| l || c | c | c | c | c | c | c | c | c | c | }
       \hline
       Tasks
         & \multicolumn{2}{ c |}{Surface Information} 
         & \multicolumn{3}{ c |}{Syntatic Information}
         & \multicolumn{5}{ c |}{Semantic Information} \\
         & SentLen & WC & TreeDepth & TopConst & BShift & Tense & SubjNum & ObjNum & SOMO & CoordInv \\
       \hline
       \hline
       \multicolumn{11}{|l|}{\textit{Word Embeddings}} \\
       \hline
       Bov-fastText (\cite{Conneau2018}) &
       54.8 & 91.6 & 32.3 & 63.1 & 50.8 & 87.8 & 81.9 & 79.3 & 50.3 & 52.7 \\
       Our model ($g_{emb}$) - Max &
       62.4 & 43.0 & 32.5 & 76.3 & 74.5 & 88.1 & 85.7 & 82.7 & 54.7 & 56.9 \\
       Our model ($g_{emb}$) - Average &
       72.1 & 70.0 & 38.5 & 79.9 & 81.4 & 89.7 & 88.5 & 86.5 & 57.4 & 63.0 \\       
       \hline
       \hline
       \multicolumn{11}{|l|}{\textit{BiLSTM-max encoders}} \\
       \hline
       SkipThought (\citeauthor{Conneau2018}) & 
       59.6 & 35.7 & 42.7 & 70.5 & 73.4 & 90.1 & 83.3 & 79.0 & 70.3 & 70.1 \\
       Our model (Encoder NER $g_{ner}$) &
       50.7 & 3.24 & 19.5 & 34.2 & 57.2 & 66.6 & 63.5 & 61.6 & 50.7 & 52.0 \\
       Our model (Encoder EMD $g_{emd}$) &
       43.3 & 1.8 & 19.3 & 30.0 & 56.3 & 64.0 & 60.1 & 57.9 & 51.3 & 50.4 \\ 
       Our model (Encoder RE $g_{re}$) &
       56.8 & 1.2 & 19.3 & 24.5 & 53.9 & 62.3 & 60.8 & 57.1 & 50.4 & 52.2 \\ 
       Our model (Encoder CR $g_{cr}$) &
       61.9 & 11.0 & 29.5 & 55.9 & 70.0 & 82.8 & 83.0 & 76.5 & 53.3 & 58.7 \\
       \hline
    \end{tabular}%
}
\end{table*}

\subsection{Multi-task learning accelerating training}

It is also interesting to understand the influence of a multi-task learning framework on the training time of the model. In the following section, we compare the speed of training in terms of number of parameter updates (a parameter update being equal to a back-propagation pass) for each of the tasks in the multi-task framework against a single-task framework. The speed of training is defined as the number of updates necessary to reach convergence based on the validation metric.

Results are presented in Table~\ref{table:speed_of_training} for the best performing multi-task model (A-GM). The multi-task framework needs less updates to reach comparable (or higher) $F_{1}$ score in most cases except on the RE task. This supports the intuition that knowledge gathered from one task is beneficial to the other tasks in the hierarchical architecture of our model.

\begin{table}
\centering
\caption{Speed of training: Difference in number of updates necessary before convergence: Multi-task (Full Model: A-GM) compared to single task. We report the differences in $F_{1}$ performance. Lower time is better, higher performance is better.}
\label{table:speed_of_training}
\resizebox{0.8\columnwidth}{!}{%
    \begin{tabular}{|l l|| c c |}
       \hline
        Setup & Model & Time $\Delta$ & Performance $\Delta$\\
        \hline
        (B) & NER & -16\% & -0.02 \\
        (C) & EMD & -44\% & +1.14 \\
        (D) & RE & +78\% & +6.76\\
        (E-GM) & Coref-GM & -28\% & +0.91\\
        \hline
    \end{tabular}%
}
\end{table}

\section{Related work}

Our work is most related to \citeauthor{Hashimoto2017} \shortcite{Hashimoto2017} who develop a joint hierarchical model trained on syntactic and semantic tasks. The top layers of this model are supervised by semantic relatedness and textual entailment between two input sentences, implying that the lower layers need to be run two times on different input sentences. Our choice of tasks avoids such a setup. Our work also adopts a different approach to multi-task training (proportional sampling) therefore avoiding the use of complex regularization schemes to prevent catastrophic forgetting. Our results show that strong performances can be reached without these ingredients. In addition, the tasks examined in this work are more focused on learning semantic representations, thereby reducing the need to learn surface and syntactic information, as evidenced by the linguistic probing tasks.

Unlike \cite{Hashimoto2017} and other previous work \cite{Katiyar2017,Bekoulis2018,Augenstein2018}, we do not learn label embeddings, meaning that the (supervised) output/prediction of a layer is not directly fed to the following layer through an embedding learned during training. Nonetheless, sharing embeddings and stacking hierarchical encoders allows us to share the supervision from each task along the full structure of our model and achieve state-of-the-art performance.

Unlike some studies on multi-task learning such as \cite{Subramanian2018}, each task has its own contextualized encoder (multi-layer BiLSTM) rather than a shared one, which we found to improve the performance. 

Our work is also related to \citeauthor{Sogaard2016} \shortcite{Sogaard2016} who propose to cast a cascade architecture into a multi-task learning framework. However, this work was focused on syntactic tasks and concluded that adding a semantic task like NER to a set of syntactic tasks does not bring any improvement. This confirms the intuition that multi-task learning is mostly effective when tasks are related and that syntactic and semantic tasks may be too distant to take advantage of shared representations. The authors used a linear classifier with a softmax activation, relying on the richness on the embeddings. As a consequence the sequence tagging decisions are made independently for each token, in contrast to our work.

One central question in multi-task learning is the training procedure. Several schemes have been proposed in the literature. \citeauthor{Hashimoto2017} \shortcite{Hashimoto2017} train their hierarchical model following the model's architecture from bottom to top: the trainer successively goes through the whole dataset for each task before moving to the task of the following level. The underlying hypothesis is that the model should perform well on low-level tasks before being trained in more complicated tasks. \citeauthor{Hashimoto2017} avoid catastrophic forgetting by introducing \textit{successive regularization} using slack constraints on the parameters. \citeauthor{Subramanian2018} \shortcite{Subramanian2018} adopt a simpler strategy for each parameter update: a task is randomly selected and a batch of the associated dataset is sampled for the current update. More recently, \citeauthor{Mccann2018} \shortcite{Mccann2018} explored various batch-level sampling strategies and showed that an anti-curriculum learning strategy \cite{Bengio2009} is most effective. In contrast, we propose a novel proportional sampling strategy, which we find to be more effective.

Regarding the selection of the set of tasks, our work is closest to \cite{Durrett2014,Singh2013}. \citeauthor{Durrett2014} \shortcite{Durrett2014} combine coreference resolution, entity linking (sometimes referred to as \textit{Wikification}) and mention detection. \citeauthor{Singh2013} \shortcite{Singh2013} combine entity tagging, coreference resolution and relation extraction. These two works are based on graphical models with hand-engineered factors. We are using a neural-net-based approach fully trainable in an end-to-end fashion, with no need for external NLP tools (such as in \cite{Durrett2014}) or hand-engineered features. Coreference resolution is rarely used in combination with other tasks. The main work we are aware of is \cite{Dhingra2018}, which uses coreference clusters to improve reading comprehension and the works on language modeling by \citeauthor{Ji2017} \shortcite{Ji2017} and \citeauthor{Yang2016lm} \shortcite{Yang2016lm}.

Regarding the combination of entity mention detection and relation, we refer to our baselines detailed above. Here again, our predictors do not require additional features like dependency trees \cite{Miwa2016} or hand-engineered heuristics \cite{Li2014}.

\section{Conclusion}

We proposed a hierarchically supervised multi-task learning model focused on a set of semantic task. This model achieved state-of-the-art results on the tasks of Named Entity Recognition, Entity Mention Detection and Relation Extraction and competitive results on Coreference Resolution while using simpler training and regularization procedures than previous works. The tasks share common embeddings and encoders allowing an easy information flow from the lowest level to the top of the architecture. We analyzed several aspects of the representations learned by this model as well as the effect of each tasks on the overall performances of the model.

\section{Acknowledgments}

We thank Gianis Bekoulis and Victor Quach for helpful feedbacks and the anonymous reviewers for constructive comments on the paper.

\fontsize{9.0pt}{10.0pt} \selectfont
\bibliography{mybib} 
\bibliographystyle{aaai}

\end{document}